%% file: main.tex
\documentclass[11pt,a4paper]{article}
\usepackage[hyperref]{emnlp2020}
\usepackage{times}
\usepackage{latexsym}

\usepackage{microtype}
\input{mydef}

\usepackage{trackchanges}
\addeditor{YJ}
\addeditor{Hanjie}

\usepackage{cleveref}
\usepackage{bbm}
\usepackage{subfigure}
\usepackage{xcolor}
\usepackage{soul}
\usepackage{xspace}

\newcommand{\hlc}[2][yellow]{{%
		\colorlet{foo}{#1}%
		\sethlcolor{foo}\hl{#2}}%
}

\crefformat{section}{\S#2#1#3}
\crefformat{subsection}{\S#2#1#3}
\crefformat{subsubsection}{\S#2#1#3}

\aclfinalcopy 
\def\aclpaperid{2585} 

\newcommand{\ourmethod}{\textsc{Vmask}\xspace}

\title{Learning Variational Word Masks to Improve the Interpretability of Neural Text Classifiers}

\author{Hanjie Chen, Yangfeng Ji\\
	Department of Computer Science \\
	University of Virginia \\
	Charlottesville, VA, USA \\
	\texttt{\{hc9mx, yangfeng\}@virginia.edu} \\}

\date{}

\begin{document}
\maketitle

\begin{abstract}
	\input{abstract}
\end{abstract}

\input{intro}

\input{relate}

\input{method}

\input{exp}

\input{con}

\section*{Acknowledgments}

We thank the authors of \citet{schulz2020restricting} and \citet{chen2018learning} for sharing their high-quality code. We thank the anonymous reviewers for many valuable comments.

\bibliographystyle{acl_natbib}
\bibliography{ref}

\clearpage
\newpage
\appendix
\input{appendix}

\end{document}

%% file: mydef.tex
\usepackage{amsmath,amsfonts,amssymb,bm}
\usepackage{pifont} 
\usepackage{graphicx,subfigure,epsfig,fancybox} 
\usepackage{float}
\usepackage{color} 
\usepackage{multirow}
\usepackage{natbib}

\usepackage{tikz}
\usetikzlibrary{shapes.geometric, positioning, calc}
\usetikzlibrary{arrows,shapes,calc}
\usetikzlibrary{trees,positioning,mindmap,shadows,fit}
\usetikzlibrary{decorations.pathreplacing}
\usetikzlibrary{tikzmark} 
\usetikzlibrary{intersections} 

\newcommand{\revised}[1]{\textcolor{blue}{}}
\renewcommand{\vec}[1]{\boldsymbol{#1}}

\newcommand{\expect}[2]{\mathbb{E}_{#2}[{#1}]}

\newcommand{\vth}[0]{\vec{\theta}}
\newcommand{\vphi}{\vec{\phi}}

\newcommand{\kldist}[2]{\text{KL}[#1\| #2]}

\usepackage{adjustbox}
\usepackage{array}
\usepackage{booktabs}

\newcolumntype{R}[2]{%
    >{\adjustbox{angle=#1,lap=\width-(#2)}\bgroup}%
    l%
    <{\egroup}%
}

\usepackage{setspace}


%% file: abstract.tex


To build an interpretable neural text classifier, most of the prior work has focused on designing inherently interpretable models or finding faithful explanations.
A new line of work on improving model interpretability has just started, and many existing methods require either prior information or human annotations as additional inputs in training.
To address this limitation, we propose the variational word mask (\ourmethod) method to automatically learn task-specific important words and reduce irrelevant information on classification, which ultimately improves the interpretability of model predictions. 
The proposed method is evaluated with three neural text classifiers (CNN, LSTM, and BERT) on seven benchmark text classification datasets. 
Experiments show the effectiveness of \ourmethod in improving both model prediction accuracy and interpretability.

%% file: intro.tex
\section{Introduction}
\label{sec:intro}

Neural network models have achieved remarkable performance on text classification due to their capacity of representation learning on natural language texts \citep{zhang2015character,yang2016hierarchical,joulin2017bag, devlin2018bert}.
However, the lack of understanding of their prediction behaviors has become a critical issue for reliability and trustworthiness and hindered their applications in the real world \citep{lipton2016mythos, ribeiro2016should, jacovi2020towards}.
Many explanation methods have been proposed to provide post-hoc explanations for neural networks \citep{ribeiro2016should, lundberg2017unified, sundararajan2017axiomatic}, but they are only able to explain model predictions and cannot help improve their interpretability. 

\input{tables/tab-intro-examples}

In this work, we consider interpretability as an intrinsic property of neural network models. Furthermore, we hypothesize that neural network models with similar network architectures could have different levels of interpretability, even though they may have similar prediction performance.
\autoref{tab:intro_examples} shows explanations extracted from two neural text classifiers with similar network architectures.\footnote{The similarity will be detailed in \autoref{sec:exp} and more examples are provided in \autoref{tab:examples}.}
Although both models make correct predictions of the sentiment polarities of two input texts (positive for example 1 and negative for example 2), they have different explanations for their predictions.
In both examples, no matter which explanation generation method is used, explanations from model B are easier to be interpreted regarding the corresponding predictions.
Motivated by the difference of interpretability, we would like to investigate the possibility of \emph{building more interpretable neural classifiers with a simple modification on input layers}.
The proposed method does not demand significant efforts on engineering network architectures \citep{rudin2019stop,melis2018towards}.
Also, unlike prior work on improving model interpretability \citep{erion2019learning,plumb2019regularizing}, it does not require pre-defined important attributions or pre-collected explanations.

Specifically, we propose variational word masks (\ourmethod) that are inserted into a neural text classifier, after the word embedding layer, and trained jointly with the model. \ourmethod learns to restrict the information of globally irrelevant or noisy word-level features flowing to subsequent network layers, hence forcing the model to focus on important features to make predictions.
Experiments in \autoref{sec:results} show that this method can improve model interpretability and prediction performance. 
As \ourmethod is deployed on top of the word-embedding layer and the major network structure keeps unchanged, it is \emph{model-agnostic} and can be applied to any neural text classifiers.

The contribution of this work is three-fold:
(1) we proposed the \ourmethod method to learn global task-specific important features that can improve both model interpretability and prediction accuracy;
(2) we formulated the problem in the framework of information bottleneck (IB) \citep{tishby2000information, tishby2015deep} and derived a lower bound of the objective function via the variational IB method \citep{alemi2016deep};
and (3) we evaluated the proposed method with three neural network models, CNN~\cite{kim2014convolutional}, LSTM~\cite{hochreiter1997long}, and BERT~\cite{devlin2018bert}, on seven text classification tasks via both quantitative and qualitative evaluations.

Our implementation is available at \url{https://github.com/UVa-NLP/VMASK}.

%% file: tables/tab-intro-examples.tex
\begin{table}
  \small
  \centering
  \begin{tabular}{p{0.01\textwidth}p{0.03\textwidth}p{0.35\textwidth}}
    \toprule
    Ex. & Model & ~~~~~~Text \& Explanation \\
    \midrule
     \multirow{2}{*}{~1} & ~~~A & An \hlc[pink!40]{exceedingly} clever \hlc[pink!60]{piece} of \hlc[pink!100]{cinema} \\
     \rule{0pt}{3ex} 
     & ~~~B & An \hlc[cyan!40]{exceedingly} \hlc[cyan!60]{clever} \hlc[cyan!20]{piece} of cinema \\
     \midrule
     \multirow{2}{*}{~2} & ~~~A & It \hlc[pink!40]{becomes} gimmicky \hlc[pink!100]{instead} \hlc[pink!60]{of} compelling \\
     \rule{0pt}{3ex} 
     & ~~~B & It \hlc[cyan!20]{becomes} \hlc[cyan!60]{gimmicky} \hlc[cyan!40]{instead} of compelling \\
    \bottomrule
  \end{tabular}
  \caption{Model A and B are two neural text classifiers with similar network architectures. They all make correct sentiment predictions on both texts (ex. 1: positive; ex. 2: negative). Two post-hoc explanation methods, LIME \citep{ribeiro2016should} and SampleShapley \citep{kononenko2010efficient}, are used to explain the model predictions on example 1 and 2 respectively. Top three important words are shown in pink or blue for model A and B.
    Whichever post-hoc method is used, explanations from model B are easier to understand because the sentiment keywords ``clever" and ``gimmicky" are highlighted.}
  \label{tab:intro_examples}
\end{table}

%% file: relate.tex
\section{Related Work}
\label{sec:relate}
Various approaches have been proposed to interpret DNNs, ranging from designing inherently interpretable models \citep{melis2018towards, rudin2019stop}, to tracking the inner-workings of neural networks \citep{jacovi2018understanding, murdoch2018beyond}, to generating post-hoc explanations \citep{ribeiro2016should, lundberg2017unified}. 
Beyond interpreting model predictions, the explanation generation methods are also promising in improving model's performance. We propose an information-theoretic method to improve both prediction accuracy and interpretability.

\paragraph{Explanation from the information-theoretic perspective.}
A line of works that motivate ours leverage information theory to produce explanations, either maximizing mutual information to recognize important features \citep{chen2018learning, guan2019towards}, or optimizing the information bottleneck to identify feature attributions \citep{schulz2020restricting, bang2019explaining}. The information-theoretic approaches are efficient and flexible in identifying important features. Different from generating post-hoc explanations for well-trained models, we utilize information bottleneck to train a more interpretable model with better prediction performance.

\paragraph{Improving prediction performance via explanations.}
Human-annotated explanations have been utilized to help improve model prediction accuracy \citep{zhang2016rationale}.
Recent work has been using post-hoc explanations to regularize models on prediction behaviors and force them to emphasize more on predefined important features, hence improving their performance \citep{ross2017right, ross2018improving, liu2019incorporating, rieger2019interpretations}.
Different from these methods that require expert prior information or human annotations, the \ourmethod method learns global important features automatically during training and incorporate them seamlessly on improving model prediction behaviors.

\paragraph{Improving interpretability via explanations.}
Some work focuses on improving model's interpretability by aligning explanations with human-judgements \citep{camburu2018snli, du2019learning, chen2019improving, erion2019learning, plumb2019regularizing}. Similarly to the prior work on improving model prediction performance, these methods still rely on annotations or external resources. Although enhancing model interpretability, they may cause the performance drop on prediction accuracy due to the inconsistency between human recognition and model reasoning process \citep{jacovi2020towards}. Our approach can improve both prediction accuracy and interpretability without resorting to human-judgements.

%% file: method.tex
\section{Method}
\label{sec:method}

This section introduces the proposed \textsc{Vmask} method.
For a given neural text classifier, the only modification on the neural network architecture is to insert a word mask layer between the input layer (e.g., word embeddings) and the representation learning layer.
We formulate our idea within the information bottleneck framework~\citep{tishby2000information}, where the word mask layer restricts the information from words to the final prediction.

\subsection{Interpretable Text Classifier with Word Masks}
\label{subsec:inter_NN}

For an input text $\vec{x}=[x_{1}, \cdots, x_{T}]$, where $x_t$ ($t\in \{1,\dots, T\}$) indicates the word or the word index in a predefined vocabulary.
In addition, we use $\vec{x}_{t}\in \mathbb{R}^{d}$ as the word embedding of $x_t$.
A neural text classifier is denoted as $f_{\vth}(\cdot)$ with parameter $\vth$, which by default takes $\vec{x}$ as input and generates a probability of output $\vec{Y}$, $p(\vec{Y}|\vec{x})$, over all possible class labels.
In this work, beyond prediction accuracy, we also expect the neural network model to be more interpretable, by focusing on important words to make predictions.

To help neural network models for better feature selection, we add a random layer $\vec{R}$ after the word embeddings, where $\vec{R}=[R_{x_1},\dots,R_{x_T}]$ has the same length of $\vec{x}$.
Each $R_{x_t}\in\{0,1\}$ is a binary random variable associated with the word type $x_t$ instead of the word position. 
This random layer together with word embeddings form the input to the neural network model, i.e., 
\begin{equation}
  \label{eq:mask_product}
  \vec{Z}=\vec{R}\odot \vec{x},
\end{equation}
where $\odot$ is an element-wise multiplication and each $\vec{Z}_{t} = R_{x_t}\cdot \vec{x}_{t}$.
Intuitively, $\vec{Z}$ only contains a subset of $\vec{x}$, which is selected randomly by $\vec{R}$.
Since $\vec{R}$ is applied directly on the words as a sequence of 0-1 masks, we also call it the word mask layer in this work. 

To ensure $\vec{Z}$ has enough information on predicting $\vec{Y}$ while contains the least redundant information from $\vec{x}$, we follow the standard practice in the information bottleneck theory \citep{tishby2000information}, and write the objective function as
\begin{equation}
\label{eq:ob_func}
\max_{\vec{Z}} I(\vec{Z};\vec{Y}) - \beta\cdot I(\vec{Z};\vec{X}),
\end{equation}
where $\vec{X}$ as a random variable representing a generic word sequence as input, $\vec{Y}$ is the one-hot output random variable, $I(\cdot;\cdot)$ is the mutual information, and $\beta\in\mathbb{R}_{+}$ is a coefficient to balance the two mutual information items.
This formulation reflects our exact expectation on $\vec{Z}$.
The main challenge here is to compute the mutual information.

\subsection{Variational Word Masks}
\label{subsec:global_smask}
Inspired by the variational information bottleneck proposed by \citet{alemi2016deep}, instead of computing $p(\vec{X},\vec{Y},\vec{Z})$, we start from an approximation distribution $q(\vec{X},\vec{Y},\vec{Z})$.
Then, with a few assumptions specified in the following, we construct a tractable lower bound of the objective in \autoref{eq:ob_func} and the detailed derivation is provided in \autoref{sec:proof}.


For $I(\vec{Z};\vec{Y})$ under $q$, we have $I(\vec{Z};\vec{Y})=\sum_{\vec{y}, \vec{z}}q(\vec{y},\vec{z})\log (q(\vec{y}|\vec{z})/q(\vec{y}))$.
By replacing $\log q(\vec{y}|\vec{z})$ with the conditional probability derived from the true distribution $\log p(\vec{y}|\vec{z})$, we introduce the constraint between $\vec{Y}$ and $\vec{Z}$ from the distribution and also 
obtain a lower bound of $I(\vec{Z};\vec{Y})$,
\begin{eqnarray}
    \label{eq:izy_bound}
    I(\vec{Z};\vec{Y})
    &\geq&\sum_{\vec{y}, \vec{z}} q(\vec{y},\vec{z}) \log p(\vec{y}|\vec{z}) + H_q(\vec{Y})\nonumber\\
    &=& \sum_{\vec{y}, \vec{z}, \vec{x}} q(\vec{x}, \vec{y})q(\vec{z}|\vec{x}) \log p(\vec{y}|\vec{z})\nonumber\\
    & & +H_q(\vec{Y}),
\end{eqnarray}
where $H_q(\cdot)$ is entropy, and the last step uses $q(\vec{x},\vec{y},\vec{z}) = q(\vec{x})q(\vec{y}|\vec{x}) q(\vec{z}|\vec{x})$, which is a factorization based on the conditional dependency \footnote{$\vec{Y} \leftrightarrow \vec{X} \leftrightarrow \vec{Z}$: $\vec{Y}$ and $\vec{Z}$ are independent given $\vec{X}$.}. 

Given a specific observation $(\vec{x}^{(i)},\vec{y}^{(i)})$, we define the empirical distribution $q(\vec{X}^{(i)},\vec{Y}^{(i)})$ as a multiplication of two Delta functions $q(\vec{X}^{(i)}=\vec{x}^{(i)},\vec{Y}^{(i)}=\vec{y}^{(i)}) = \delta_{\vec{x}^{(i)}}(\vec{x})\cdot\delta_{\vec{y}^{(i)}}(\vec{y})$.
Then, \autoref{eq:izy_bound} can be further simplified as
\begin{eqnarray}
    \label{eq:izy_bound_1}
    I(\vec{Z};\vec{Y}^{(i)})
    &\geq& \sum_{\vec{z}}q(\vec{z}|\vec{x}^{(i)}) \log p(\vec{y}^{(i)}|\vec{z})\nonumber\\
    &=& \expect{\log p(\vec{y}^{(i)}|\vec{z})}{q(\vec{z}|\vec{x}^{(i)})}.
\end{eqnarray}

Similarly, for $I(\vec{Z};\vec{X})$ under $q$, we have an upper bound of $I(\vec{Z};\vec{X})$ by replacing $p(\vec{Z}|\vec{X})$ with a predefined prior distribution $p_0(\vec{Z})$
\begin{eqnarray}
  \label{eq:izx_bound}
  I(\vec{Z};\vec{X})
  &\leq& \expect{\kldist{q(\vec{z}|\vec{x})}{p_0(\vec{z})}}{q(\vec{x})}\nonumber\\
  &=& \kldist{q(\vec{z}|\vec{x}^{(i)})}{p_0(\vec{z})},
\end{eqnarray}
where $\kldist{\cdot}{\cdot}$ denotes Kullback-Leibler divergence.
The simplification in the last step is similar to \autoref{eq:izy_bound_1} with the empirical distribution $q(\vec{X}^{(i)})$.

Substituting (\ref{eq:izx_bound}) and (\ref{eq:izy_bound_1}) into \autoref{eq:ob_func} gives us a lower bound $\mathcal{L}$ of the informaiton bottleneck
\begin{equation}
  \begin{aligned}
    \label{eq:ob_func_1}
    \mathcal{L}=
    &\expect{\log p(\vec{y}^{(i)}|\vec{z})}{q(\vec{z}|\vec{x}^{(i)})}\\
    &-\beta\cdot\kldist{q(\vec{z}|\vec{x}^{(i)})}{p_0(\vec{z})}.
  \end{aligned}
\end{equation}
The learning objective is to maximize \autoref{eq:ob_func_1} with respect to the approximation distribution $q(\vec{X},\vec{Y},\vec{Z})=q(\vec{X},\vec{Y})q(\vec{Z}|\vec{X})$.
As a classification problem, $\vec{X}$ and $\vec{Y}$ are both observed and $q(\vec{X},\vec{Y})$ has already been simplified as an empirical distribution, the only one left in the approximation distribution is $q(\vec{Z}|\vec{X})$.
Similarly to the objective function in variational inference \citep{alemi2016deep, rezende2015variational}, the first term in $\mathcal{L}$ is to make sure the information in $q(\vec{Z}|\vec{X})$ for predicting $\vec{Y}$, while the second term in $\mathcal{L}$ is to regularize $q(\vec{Z}|\vec{X})$ with a predefined prior distribution $p_0(\vec{Z})$.

The last step of obtaining a practical objective function is to notice that, given $\vec{X}_t^{(i)}=\vec{x}_t^{(i)}$ every $\vec{Z}_t$ can be redefined as
\begin{equation}
\vec{Z}_t = R_{x_t} \cdot \vec{x}_t^{(i)}, 
\end{equation}
where $R_{x_t} \in\{0,1\}$ is a standard Bernoulli distribution. Then, $\vec{Z}$ can be reparameterized as $\vec{Z}=\vec{R}\odot\vec{x}^{(i)}$ with $\vec{R}=[R_{x_1},\dots,R_{x_T}]$.
The lower bound $\mathcal{L}$ can be rewritten with the random variable $\vec{R}$ as
\begin{equation}
  \label{eq:ob_func_2}
  \begin{split}
    \mathcal{L}
    = & \expect{\log p(\vec{y}^{(i)}|\vec{R},\vec{x}^{(i)})}{q(\vec{r}|\vec{x}^{(i)})}\\
    & -\beta\cdot \kldist{q(\vec{R}|\vec{x}^{(i)})}{p_0(\vec{R})}.
  \end{split}
\end{equation}
Note that, although $\beta$ is inherited from the information bottleneck theory, in practice it will be used as a tunable hyper-parameter to address the notorious posterior collapse issue \citep{bowman2016generating,kim2018tutorial}.

\subsection{Connections}
\label{subsec:connection}
The idea of modifying word embeddings with the information bottleneck method has recently shown some interesting applications in NLP.
For example, \citet{li2019specializing} proposed two ways to transform word embeddings into new representations for better POS tagging and syntactic parsing.
According to \autoref{eq:mask_product}, \ourmethod can be viewed as a simple linear transformation on word embeddings.
The difference is that $\{R_{x_t}\}$ is defined on the vocabulary, therefore can be used to represent the global importance of word $x_t$.
Recall that $R_{x_t}\in\{0,1\}$, from a slightly different perspective, \autoref{eq:mask_product} can be viewed as a generalized method on word-embedding dropout \citep{gal2016theoretically}.
Although there are two major differences: (1) in \citet{gal2016theoretically} all words share the same dropout rate, while in \ourmethod every word has its own dropout rate specified by $q(R_{x_t}|\vec{x}_t)$, i.e. $1-\expect{q(R_{x_t}|\vec{x}_t)}{}$; (2) the motivation of word-embedding dropout is to force a model not to rely on single words for prediction, while $\ourmethod$ is to learn a task-specific importance for every word. 

Another implementation for making word masks sparse is by adding $L_{0}$ regularization \citep{lei2016rationalizing, bastings2019interpretable, cao2020decisions}, while in the objective \autoref{eq:ob_func_2}, we regularize masks with a predefined prior distribution $p_0(\vec{R})$ as described in \autoref{subsec:sepci_train}.

\subsection{Model Specification and Training}
\label{subsec:sepci_train}

We resort to mean-field approximation \citep{blei2017variational} to simplify the assumption on our $q$ distribution. For $q_{\vphi}(\vec{R}|\vec{x})$, we have $q_{\vphi}(\vec{R}|\vec{x}) = \prod_{t=1}^{T}q_{\phi}(R_{x_t}|\vec{x}_{t})$, which means the random variables are mutually independent and each governed by $\vec{x}_{t}$.
We use the amortized variational inference \citep{rezende2015variational} to represent the posterior distribution $q_{\vphi}(R_{x_t}|\vec{x}_t)$ with using an inference network \citep{kingma2013auto}.
In this work, we adopt a single-layer feedforward neural network as the inference network, whose parameters $\vphi$ are optimized with the model parameters $\vth$ during training. 

Following the same factorization as in $q_{\vphi}(\vec{R}|\vec{x})$, we define the prior distribution $p_0(\vec{R})$ as $p_0(\vec{R}) = \prod_{t=1}^{T}p_0(R_{x_t})$ and each of them as $p_0(R_{x_t})=\text{Bernoulli}(0.5)$. By choosing this non-informative prior, it means every word is initialized with no preference to be important or unimportant, and thus has the equal probability to be masked or selected. As $p_0(\vec{R})$ is a uniform distribution, we can further simplify the second term in \autoref{eq:ob_func_2} as a conditional entropy,
\begin{equation}
\label{eq:opti_func_3}
\max_{\vth, \vphi}\expect{\log p(\vec{y}^{(i)}|\vec{R},\vec{x}^{(i)})}{q}+\beta\cdot H_q(\vec{R}|\vec{x}^{(i)}).
\end{equation}

We apply stochastic gradient descent to solve the optimization problem (\autoref{eq:opti_func_3}). Particularly in each iteration, the first term in \autoref{eq:opti_func_3} is approximated with a single sample from $q(\vec{R}|\vec{x}^{(i)})$ \citep{kingma2013auto}. However, sampling from a Bernoulli distribution (like from any other discrete distributions) causes difficulty in backpropagation.
We adopt the Gumbel-softmax trick \citep{jang2016categorical, maddison2016concrete} to utilize a continuous differentiable approximation and tackle the discreteness of sampling from Bernoulli distributions (\autoref{sec:sampling}).
During training, We use Adam \citep{kingma2014adam} for optimization and KL cost annealing \citep{bowman2016generating} to avoid posterior collapse.

For a given word $x_t$ and its word embedding $\vec{x}_t$, in training stage, the model samples each $r_{x_t}$ from $q(R_{x_t}|\vec{x}_t)$ to decide to either keep or zero out the corresponding word embedding $\vec{x}_t$.
In inference stage, the model takes the multiplication of the word embedding $\vec{x}_t$ and the expectation of the word mask distribution, i.e. $\vec{x}_{t}\cdot\expect{q(R_{x_t}|\vec{x}_t)}{}$, as input.

%% file: exp.tex
\section{Experiment Setup}
\label{sec:exp}
The proposed method is evaluated on seven text classification tasks, ranging from sentiment analysis to topic classification, with three typical neural network models, a long short-term memories~\citep[LSTM]{hochreiter1997long}, a convolutional neural network~\citep[CNN]{kim2014convolutional}, and BERT~\citep{devlin2018bert}.

\paragraph{Datasets.} We adopt seven benchmark datasets: movie reviews IMDB \citep{maas2011learning}, Stanford Sentiment Treebank with fine-grained labels SST-1 and its binary version SST-2 \citep{socher2013recursive}, Yelp reviews \citep{zhang2015character}, AG’s News \citep{zhang2015character}, 6-class question classification TREC \citep{li2002learning}, and subjective/objective classification Subj \citep{pang2005seeing}. For the datasets (e.g. IMDB, Subj) without standard train/dev/test split, we hold out a proportion of training examples as the development set. \autoref{tab:datasets} shows the statistics of the datasets.

\input{tables/tab-datasets}

\input{tables/tab-compare-acc}

\input{tables/tab-aopc}
 
\paragraph{Models.}
The CNN model~\cite{kim2014convolutional} contains a single convolutional layer with filter sizes ranging from 3 to 5. The LSTM~\cite{hochreiter1997long} has a single unidirectional hidden layer. Both models are initialized with 300-dimensional pretrained word embeddings~\cite{mikolov2013distributed}. We fix the embedding layer and update other parameters on different datasets to achieve the best performance respectively. 
We use the pretrained BERT-base model\footnote{\url{https://github.com/huggingface/pytorch-transformers}{}} with 12 transformer layers, 12 self-attention heads, and the hidden size of 768. We fine-tune it with different downstream tasks, and then fix the embedding layer and train the mask layer with the rest of the model together.

\paragraph{Baselines and Competitive Methods.}
As the goal of this work is to propose a novel training method that improves both prediction accuracy and interpretability, we employ two groups of models as baselines and competitive systems.
Models trained with the proposed method are named with suffix ``-\ourmethod''.
We also provide two baselines: (1) models trained by minimizing the cross-entropy loss (postfixed with ``-base'') and (2) models trained with $\ell_2$-regularization (postfixed with ``-$\ell_2$'').
The comparison with these two baseline methods mainly focuses on prediction performance as no explicit training strategies are used to improve interpretability.

Besides, we also propose two competitive methods: models trained with the explanation framework ``Learning to Explain'' \citep{chen2018learning} (postfixed with ``-L2X'') and the ``Information Bottleneck Attribution'' \citep{schulz2020restricting} (postfixed with ``-IBA'').
L2X and IBA were originally proposed to find feature attributions as post-hoc explanations for well-trained models. We integrated them in model training, working as the mask layer to directly generate mask values for input features (L2X) or restrict information flow by adding noise (IBA). In our experiments, all training methods worked with random dropout ($\rho=0.2$) to avoid overfitting.

More details about experiment setup are in \autoref{sec:sup_exp}, including data pre-processing, model configurations, and the implementation of L2X and IBA in our experiments.

\section{Results and Discussion}
\label{sec:results}

We trained the three models on the seven datasets with different training strategies. \autoref{tab:compare-acc} shows the prediction accuracy of different models on test sets. The validation performance and average runtime are in \autoref{sec:val_run}.
As shown in \autoref{tab:compare-acc}, all base models have the similar prediction performance comparing to numbers reported in prior work (\autoref{sec:pre_results}).
The models trained with \ourmethod outperform the ones with similar network architectures but trained differently.
The results show that \ourmethod can help improve the generalization power.

Except the base models and the models trained with the proposed method, the records of other three competitors are mixed.
For example, the traditional $\ell_2$-regularization cannot always help improve accuracy, especially for the BERT model.
Although the performance with IBA is slightly better than with L2X, training with them does not show a constant improvement on a model's prediction accuracy. 

To echo the purpose of improving model interpretability, the rest of this section will focus on evaluating the model interpretability quantitatively and qualitatively.

\subsection{Quantitative Evaluation}
\label{subsec:quanti_eva}
We evaluate the local interpretability of \ourmethod-based models against the base models via the AOPC score \citep{nguyen2018comparing,samek2016evaluating} and the global interpretability against the IBA-based models via post-hoc accuracy \citep{chen2018learning}.
Empirically, we observed the agreement between local and global interpretability, so there is no need to exhaust all possible combinations in our evaluation.

\subsubsection{Local interpretability: AOPC}
\label{subsubsec:local_inter}
We adopt two model-agnostic explanation methods, LIME \citep{ribeiro2016should} and SampleShapley \citep{kononenko2010efficient}, to generate local explanations for base and \ourmethod-based models, where ``local'' means explaining each test data individually. The area over the perturbation curve (AOPC)~\citep{nguyen2018comparing, samek2016evaluating} metric is utilized to evaluate the faithfulness of explanations to models. It calculates the average change of prediction probability on the predicted class over all test data by deleting top $n$ words in explanations. 
We adopt this metric to evaluate the model interpretability to post-hoc explanations. 
Higher AOPC scores are better.

For TREC and Subj datasets, we evaluate all test data. For each other dataset, we randomly pick up 1000 examples for evaluation due to  computation costs. \autoref{tab:aopc} shows the AOPCs of different models on the seven datasets by deleting top 5 words identified by LIME or SampleShapley.  The AOPCs of \ourmethod-based models are significantly higher than that of base models on most of the datasets, indicating that \ourmethod can improve model's interpretability to post-hoc explanations. The results on the TREC dataset are very close because top 5 words are possible to include all informative words for short sentences with the average length of 10.

\subsubsection{Global Interpretability: Post-hoc accuracy}
\label{subsubsec:post_hoc_acc}
The expectation values $\{\expect{q(R_{x_t}|\vec{x}_t)}{}\}$ represent the global importance of words (\autoref{subsec:connection}) for a specific task.
To measure the interpretability of a model itself (aka, global interpretability),
we adopt the post-hoc accuracy \citep{chen2018learning} to evaluate the influence of global task-specific important features on the predictions of \ourmethod- and IBA-based models. For each test data, we select the top $k$ words based on their global importance scores for the model to make a prediction, and compare it with the original prediction made on the whole input text
\begin{equation*}
  \label{eq:post-hoc-acc}
  \text{post-hoc-acc}(k)=\frac{1}{M}\sum_{m=1}^M\mathbbm{1} [y_{m}(k)=y_{m}],
\end{equation*}
where $M$ is the number of examples, $y_{m}$ is the predicted label on the m-th test data, and $y_{m}(k)$ is the predicted label based on the top $k$ important words. 

\begin{figure*}[htb]
  \centering
  \subfigure[IMDB]{
    \label{fig:imdb_post_hoc_acc}
    \includegraphics[width=0.22\textwidth]{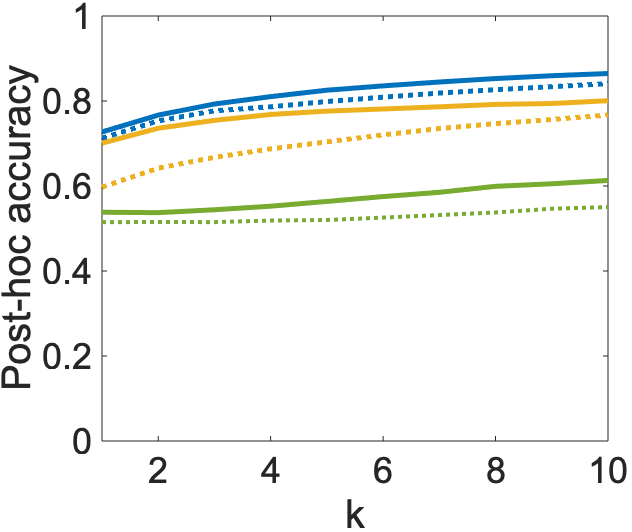}}
  \subfigure[SST-1]{
    \label{fig:sst1_post_hoc_acc}
    \includegraphics[width=0.22\textwidth]{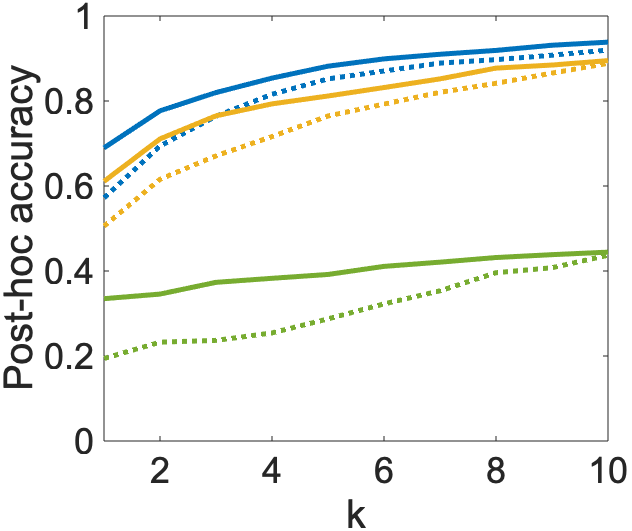}}
  \subfigure[SST-2]{
    \label{fig:sst2_post_hoc_acc}
    \includegraphics[width=0.22\textwidth]{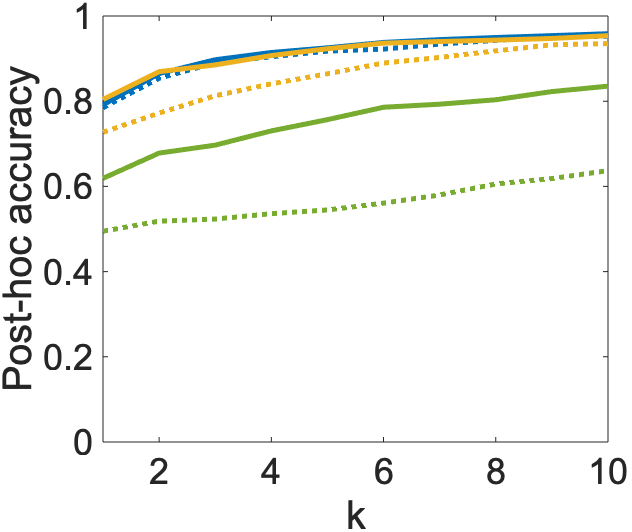}}
  \subfigure[Yelp]{
    \label{fig:yelp_post_hoc_acc}
    \includegraphics[width=0.22\textwidth]{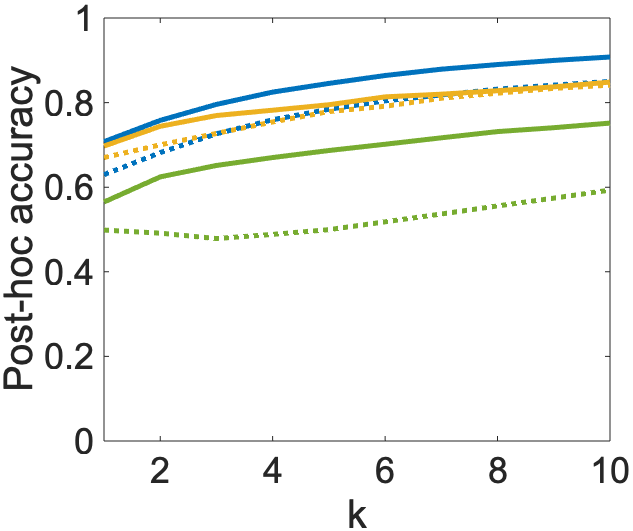}}
  \subfigure[AG News]{
    \label{fig:agnews_post_hoc_acc}
    \includegraphics[width=0.22\textwidth]{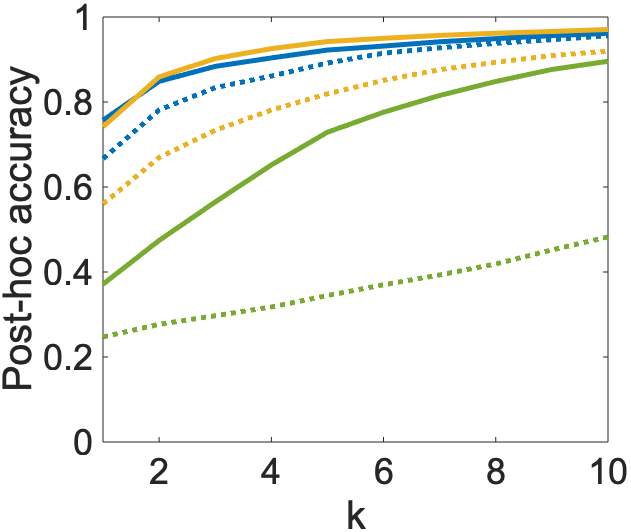}}
  \subfigure[TREC]{
    \label{fig:trec_post_hoc_acc}
    \includegraphics[width=0.22\textwidth]{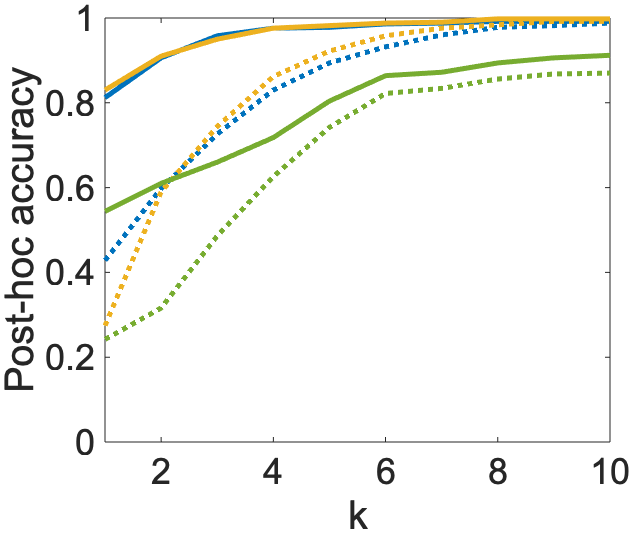}}
  \subfigure[Subj]{
    \label{fig:subj_post_hoc_acc}
    \includegraphics[width=0.32\textwidth]{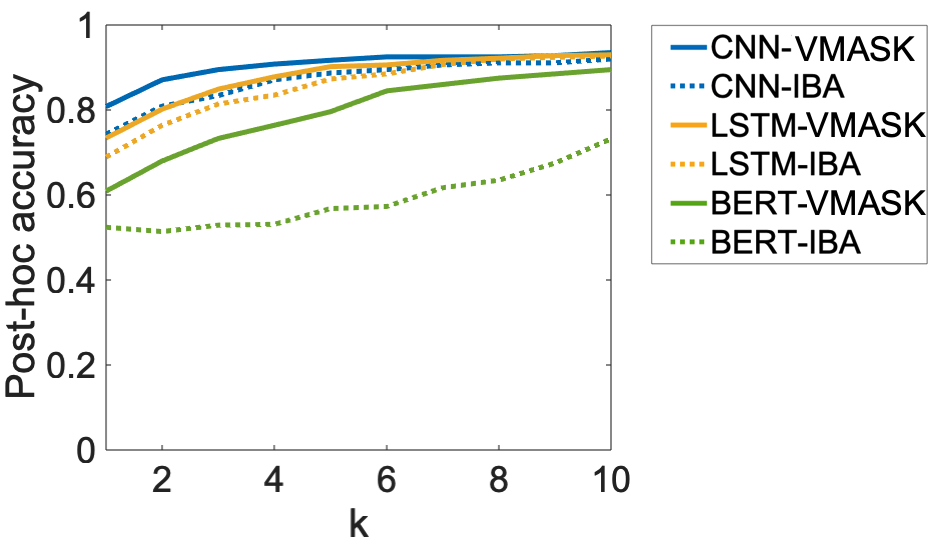}}
  \caption{Post-hoc accuracy of \ourmethod- and IBA-based models on the seven datasets.}
  \label{fig:post_hoc_acc}
\end{figure*}

\autoref{fig:post_hoc_acc} shows the results of \ourmethod- and IBA-based models on the seven datasets with $k$ ranging from 1 to 10. \ourmethod-based models (solid lines) outperform IBA-based models (dotted lines) with higher post-hoc accuracy, which indicates our proposed method is better on capturing task-specific important features. 
For CNN-\ourmethod and LSTM-\ourmethod, using only top two words can achieve about $80\%$ post-hoc accuracy, even for the IMDB dataset, which has the average sentence length of 268 tokens.
The results illustrate that \ourmethod can identify informative words for model predictions.
We also noticed that BERT-\ourmethod has lower post-hoc accuracy than the other two models.
It is probably because BERT tends to use larger context with its self-attentions for predictions.
This also explains that the post-hoc accuracies of BERT-\ourmethod on the IMDB and SST-1 datasets are catching up slowly with $k$ increasing.

\input{tables/tab-examples}

\subsection{Qualitative Evaluation}
\label{subsec:qual_eva}

\paragraph{Visualizing post-hoc local explanations.}
\autoref{tab:examples} shows some examples of LIME explanations for different models on the IMDB dataset. We highlight the top three important words identified by LIME, where the color saturation indicates word attribution. The pair of base and \ourmethod-based models make the same and correct predictions on the input texts. For \ourmethod-based models, LIME can capture the sentiment words that indicate the same sentiment polarity as the prediction. While for base models, LIME selects some irrelevant words (e.g. ``plot", ``of", ``to") as explanations, which illustrates the relatively lower interpretability of base models to post-hoc explanations.

\paragraph{Visualizing post-hoc global explanations.}
We adopt SP-LIME proposed by \citet{ribeiro2016should} as a third-party  global interpretability of base and \ourmethod-based models. Without considering the rectriction on the number of explanations, we follow the method to compute feature global importance from LIME local explanations (\autoref{subsubsec:local_inter}) by calculating the sum over all local importance scores of a feature as its global importance. To distinguish it from the global importance learned by \ourmethod, we call it \emph{post-hoc global importance}. 

\autoref{tab:sp_lime} lists the top three post-hoc global important words of base and \ourmethod-based models on the IMDB dataset. For \ourmethod-based models, the global important features selected by SP-LIME are all sentiment words. While for base models, some irrelevant words (e.g. ``performances", ``plot", ``butcher") are identified as important features, which makes model predictions unreliable.

\begin{figure}[tbh]
  \centering
  \includegraphics[width=0.45\textwidth]{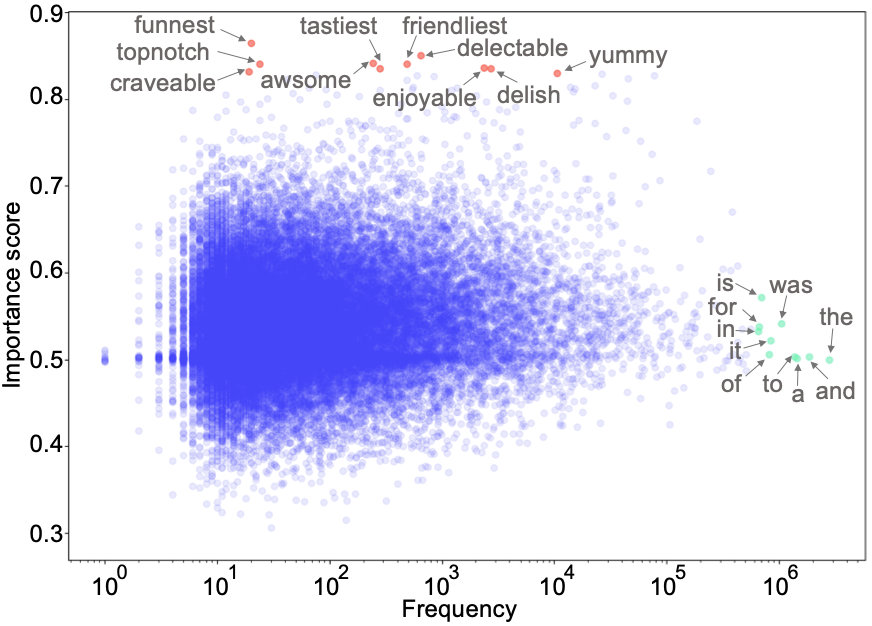}
  \caption{\label{fig:freq_imp} Scatter plot of word global importance and frequency (in log scale) of LSTM-\ourmethod on the Yelp dataset, where red dots represent top 10 important sentiment words and green dots represent top 10 high-frequency words.}
\end{figure}

\paragraph{Frequency-importance correlation.}
We compute the Pearson correlation coefficients between word frequency and global word importance of \ourmethod-based models in \autoref{sec:p_corr}.
The results show that they are not significantly correlated, which indicates that \ourmethod is not simply learning to select high-frequency words.
\autoref{fig:freq_imp} further verifies this by ploting the expectation ($\expect{q(R_{x_t}|\vec{x}_t)}{}$) of word masks from the LSTM-\ourmethod trained on Yelp and the word frequency from the same dataset.
Here, we visualize the top 10 high-frequency words and top 10 important words based the expectation of word masks.
The global importance scores of the sentiment words are over 0.8, even for some low-frequency words (e.g. ``funnest", ``craveable"), while that of the high-frequency words are all around 0.5, which means the \ourmethod-based models are less likely to focus on the irrelevant words to make predictions.

\input{tables/tab-sp-lime}
\begin{figure}[htb]
  \centering
  \subfigure[CNN-\ourmethod]{
    \label{fig:cnn_agnews_wordcloud}
    \includegraphics[width=0.2\textwidth]{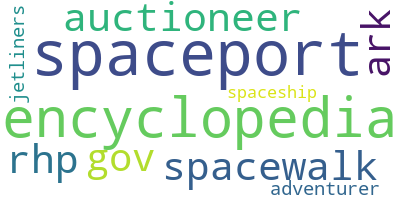}}
  \subfigure[CNN-IBA]{
    \label{fig:cnn_agnews_wordcloud_iba}
    \includegraphics[width=0.2\textwidth]{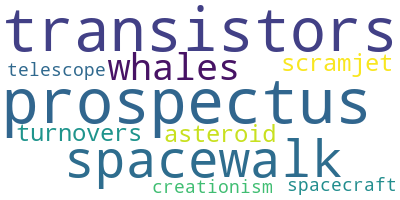}}
  \subfigure[LSTM-\ourmethod]{
    \label{fig:lstm_yelp_wordcloud}
    \includegraphics[width=0.2\textwidth]{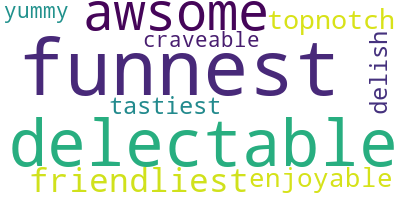}}
  \subfigure[LSTM-IBA]{
    \label{fig:lstm_yelp_wordcloud_iba}
    \includegraphics[width=0.2\textwidth]{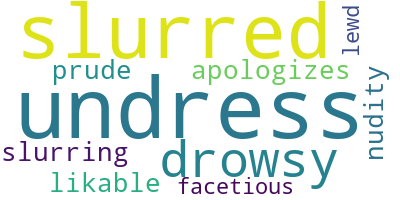}}
  \subfigure[BERT-\ourmethod]{
    \label{fig:bert_subj_wordcloud}
    \includegraphics[width=0.2\textwidth]{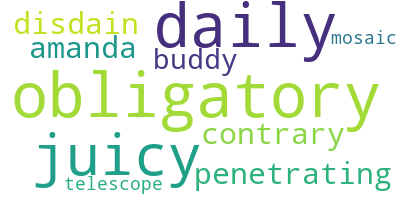}}
  \subfigure[BERT-IBA]{
    \label{fig:bert_subj_wordcloud_iba}
    \includegraphics[width=0.2\textwidth]{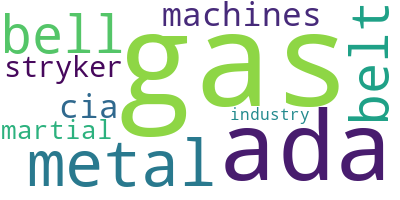}}
  \caption{Word clouds of top 10 important words, where (a) is CNN-\ourmethod on the AG News dataset, (b) is CNN-IBA on the AG News dataset, (c) is LSTM-\ourmethod on the Yelp dataset, (d) is LSTM-IBA on the Yelp dataset, (e) is BERT-\ourmethod on the Subj dataset, and (f) is BERT-IBA on the Subj dataset.}
  \label{fig:word_cloud}
\end{figure}

\paragraph{Task-specific important words.}
\autoref{fig:word_cloud} visualizes top 10 important words for the \ourmethod- and IBA-based models on three datasets via word clouds. We can see that the selected words by \ourmethod are consistent with the corresponding topic, such as ``funnest", ``awsome" for sentiment analysis, and ``encyclopedia", ``spaceport" for news classification, while IBA selects some irrelevant words (e.g. ``undress", ``slurred").

%% file: tables/tab-datasets.tex
\begin{table}
  \centering
  \begin{tabular}{llllll}
    	\toprule
         Datasets & \textit{C} & \textit{L} & \textit{\#train} & \textit{\#dev} & \textit{\#test} \\
         \midrule
         IMDB & 2 & 268 & 20K & 5K & 25K  \\
         SST-1 & 5 & 18 & 8544 & 1101 & 2210  \\
         SST-2 & 2 & 19 & 6920 & 872 & 1821  \\
         Yelp & 2 & 138 & 500K & 60K  & 38K  \\
         AG News & 4 & 32 & 114K & 6K  & 7.6K  \\
         TREC & 6 & 10 & 5000 & 452  & 500  \\
         Subj & 2 & 23 & 8000 & 1000  & 1000  \\
         \bottomrule
  \end{tabular}
  \caption{Summary statistics for the datasets, where \textit{C} is the number of classes, \textit{L} is average sentence length, and \textit{\#} counts the number of examples in the \textit{train/dev/test} sets.}
  \label{tab:datasets}
\end{table}

%% file: tables/tab-compare-acc.tex
\begin{table*}[tbph]
  \centering
  \begin{tabular}{ccccccccc}
    \toprule
    Models & Methods & IMDB & SST-1 & SST-2 & Yelp & AG News & TREC & Subj \\
    \midrule
    \multirow{5}{*}{CNN} & CNN-base & 89.06  & 46.32  & 85.50  & 94.32  & 91.30 & 92.40 & 92.80 \\
           & CNN-$\ell_2$ & 89.12  & 46.01  & 85.56  & 94.46  & 91.28 & 90.62 & 92.39 \\
           & CNN-L2X & 78.94  & 37.92  & 80.01  & 83.14  & 84.36 & 61.00 & 82.40 \\
           & CNN-IBA & 88.31  & 41.40  & 84.24  & 93.82  & 91.37 & 89.80 & 91.80 \\
           & CNN-\ourmethod & \textbf{90.10}  & \textbf{48.92}  & \textbf{85.78}  & \textbf{94.53}  & \textbf{91.60} & \textbf{93.02}  & \textbf{93.50} \\
    \midrule
    \multirow{5}{*}{LSTM} & LSTM-base & 88.39  & 43.84  & 83.74  & 95.06  & 91.03  & 90.40 & 90.20 \\
           & LSTM-$\ell_2$ & 88.40  & 43.91  & 83.36  & 95.00  & 91.09 & 90.20 & 89.10 \\
           & LSTM-L2X & 67.45  & 36.92  & 75.45  & 77.12  & 77.53 & 46.00 & 81.80 \\
           & LSTM-IBA & 88.48  & 42.99  & 83.53  & 94.74  & 91.14 & 85.40 & 89.50 \\
           & LSTM-\ourmethod & \textbf{90.07}  & \textbf{44.12}  & \textbf{84.35}  & \textbf{95.41}  & \textbf{92.19} & \textbf{90.80} & \textbf{91.20} \\
    \midrule
    \multirow{5}{*}{BERT} & BERT-base & 91.80  & 53.43  & 92.25  & 96.42  & 93.59  & 96.40 & 95.10 \\
           & BERT-$\ell_2$ & 91.75  & 52.08  & 92.25  & 96.41  & 93.52 & 96.80 & 94.80 \\
           & BERT-L2X & 71.75  & 39.23  & 74.03  & 87.14  & 82.59 & 93.20 & 86.10 \\
           & BERT-IBA & 91.66  & 53.80  & 92.24  & 96.27  & 93.45  & 96.80  & 95.60 \\
           & BERT-\ourmethod & \textbf{93.04}  & \textbf{54.53}  & \textbf{92.26}  & \textbf{96.80}  & \textbf{94.24} & \textbf{97.00} & \textbf{96.40} \\
    \bottomrule
  \end{tabular}
  \caption{Prediction accuracy (\%) of different models with different training strategies on the seven datasets.}
  \label{tab:compare-acc}
\end{table*}

%% file: tables/tab-aopc.tex
\begin{table*}[tbh]
	\centering
	\begin{tabular}{ccccccccc}
		\toprule
		Methods & Models & IMDB & SST-1 & SST-2 & Yelp & AG News & TREC & Subj \\
		\midrule
		\multirow{6}{*}{LIME} & CNN-base & 14.47 & 7.59 & 16.50 & 10.69 & 5.66 & 15.28 & 9.77  \\
		& CNN-\ourmethod  & \textbf{14.74} & \textbf{8.63} & \textbf{18.86} & \textbf{11.38} & \textbf{9.03} & 14.81 &  \textbf{12.40}  \\
		\rule{0pt}{2ex}  
		& LSTM-base  & 14.34 & 8.76 & 17.03 & 8.72 & 7.00 & 11.95 &  9.67  \\
		& LSTM-\ourmethod  & \textbf{15.10} & \textbf{9.52} & \textbf{22.14} & \textbf{9.70} & \textbf{7.39} & 11.97 &  \textbf{11.68}  \\
		\rule{0pt}{2ex}  
		& BERT-base  & 10.63 & 36.00 & 35.89 & 6.30 & 7.00 & 59.22 &  13.08  \\
		& BERT-\ourmethod  & \textbf{12.64} & 36.16 & \textbf{46.87} & 6.49 & \textbf{8.47} & \textbf{60.37} & \textbf{17.82}  \\
		\midrule
		\multirow{6}{*}{SampleShapley} & CNN-base & 15.53 & 7.63 & 13.15 & 13.57 & 9.88 & 14.97 & 8.84  \\
		& CNN-\ourmethod  & 15.53 & \textbf{8.33} & \textbf{15.95} & \textbf{15.06} & 9.98 & \textbf{15.03} &  \textbf{12.88}  \\
		\rule{0pt}{2ex}  
		& LSTM-base  & 15.80 & 7.91 & 22.38 & 10.55 & 6.62 & 11.90 &  11.66  \\
		& LSTM-\ourmethod  & \textbf{16.48} & \textbf{9.73} & 22.52 & \textbf{10.99} & \textbf{7.65} & 11.86 &  \textbf{12.74}  \\
		\rule{0pt}{2ex}  
		& BERT-base  & 12.97 & 42.06 & 43.16 & 18.06 & 7.21 & 57.69 &  33.22  \\
		& BERT-\ourmethod  & \textbf{13.18} & \textbf{44.57} & \textbf{50.44} & 18.17 & \textbf{10.02} & \textbf{58.26} & \textbf{34.22}  \\
		\bottomrule
	\end{tabular}
	\caption{AOPCs (\%) of LIME and SampleShapley in interpreting the base and \ourmethod-based models on the seven datasets.}
	\label{tab:aopc}
\end{table*}

%% file: tables/tab-examples.tex
\begin{table*}
  \centering
  \begin{tabular}{cp{0.65\textwidth}c}
    \toprule
    Models & Texts & Prediction \\
    \midrule
     CNN-base & Primary \hlc[pink!100]{plot} , primary direction , \hlc[pink!40]{poor} \hlc[pink!60]{interpretation} .  & negative \\
     CNN-\ourmethod & Primary plot , primary \hlc[cyan!20]{direction} , \hlc[cyan!60]{poor} \hlc[cyan!40]{interpretation} . & negative \\
    \rule{0pt}{3ex} 
     LSTM-base & John Leguizamo 's freak is one \hlc[pink!100]{of} the \hlc[pink!40]{funniest} one man shows I 've ever seen . I recommend it to anyone with a good sense \hlc[pink!60]{of} humor . & positive \\
     LSTM-\ourmethod & \hlc[cyan!20]{John} Leguizamo 's freak is one of the \hlc[cyan!60]{funniest} one man \hlc[cyan!40]{shows} I 've ever seen . I recommend it to anyone with a good sense of humor . & positive \\
    \rule{0pt}{3ex} 
     BERT-base & Great story , great music . A heartwarming love story that ' s beautiful \hlc[pink!100]{to} watch and delightful \hlc[pink!60]{to} listen \hlc[pink!40]{to} . Too bad there is no soundtrack CD . & positive \\
     BERT-\ourmethod & \hlc[cyan!20]{Great} story , great music . A \hlc[cyan!40]{heartwarming} love story that ' s beautiful to watch and \hlc[cyan!60]{delightful} to listen to . Too bad there is no soundtrack CD . & positive \\
    \bottomrule
  \end{tabular}
  \caption{Examples of the explanations generated by LIME for different models on the IMDB dataset, where the top three important words are highlighted. The color saturation indicates word attribution.}
  \label{tab:examples}
\end{table*}

%% file: tables/tab-sp-lime.tex
\begin{table}[tbh]
  \centering
  \begin{tabular}{cp{0.28\textwidth}}
    \toprule
    Models & Words \\
    \midrule
     CNN-base & excellent, performances, brilliant \\
     CNN-\ourmethod & excellent, fine, favorite \\
    \rule{0pt}{3ex} 
     LSTM-base & plot, excellent, liked \\
     LSTM-\ourmethod & excellent, favorite, brilliant \\
    \rule{0pt}{3ex} 
     BERT-base & live, butcher, thrilling \\
     BERT-\ourmethod & powerful, thrilling, outstanding \\
    \bottomrule
  \end{tabular}
  \caption{Post-hoc global important words selected by SP-LIME for different models on the IMDB dataset.}
  \label{tab:sp_lime}
\end{table}

%% file: con.tex
\section{Conclusion}
\label{sec:conclusion}
In this paper, we proposed an effective method, \ourmethod, learning global task-specific important features to improve both model interpretability and prediction accuracy. 
We tested \ourmethod with three different neural text classifiers on seven benchmark datasets, and assessed its effectiveness via both quantitative and qualitative evaluations.

%% file: appendix.tex
\section{Proof of the Two Bounds for the Information Bottleneck}
\label{sec:proof}

The following derivation is similar to the variational information bottleneck, where the difference is that our starting point is the approximation distribution $q(\vec{X},\vec{Y},\vec{Z})$ instead of the true distribution $p(\vec{X},\vec{Y},\vec{Z})$.

\paragraph{The lower bound for $I(\vec{Z};\vec{Y})$.}
\begin{eqnarray}
I(\vec{Z},\vec{Y})
&=& \sum_{\vec{y}, \vec{z}} q(\vec{y},\vec{z})\log \frac{q(\vec{y},\vec{z})}{q(\vec{y})q(\vec{z})}\nonumber\\
&=& \sum_{\vec{y}, \vec{z}} q(\vec{y},\vec{z})\log \frac{q(\vec{y}|\vec{z})}{q(\vec{y})}\nonumber\\
&=& \sum_{\vec{y}, \vec{z}} q(\vec{y},\vec{z}) \log q(\vec{y}|\vec{z})\nonumber\\
& & + H_q(\vec{Y}),
\end{eqnarray}
where $H_q(\cdot)$ represents entropy.
Now, if we replace $\log q(\vec{y}|\vec{z})$ with the conditional probability derived from the true distribution $\log p(\vec{y}|\vec{z})$, we have
\begin{equation}
\begin{aligned}
\label{eq:L_bound}
& \sum_{\vec{y}, \vec{z}} q(\vec{y},\vec{z})\log q(\vec{y}|\vec{z}) \\
&= \sum_{\vec{y}, \vec{z}} q(\vec{y},\vec{z})\log \frac{q(\vec{y}|\vec{z})p(\vec{y}|\vec{z})}{p(\vec{y}|\vec{z})} \\
&=\sum_{\vec{y}, \vec{z}} q(\vec{y},\vec{z})\log p(\vec{y}|\vec{z}) + \kldist{q(\vec{y}|\vec{z})}{p(\vec{y}|\vec{z})} \\
&\geq \sum_{\vec{y}, \vec{z}} q(\vec{y},\vec{z})\log p(\vec{y}|\vec{z}),
\end{aligned}
\end{equation}
where $\kldist{\cdot}{\cdot}$ denotes Kullback-Leibler divergence.
Therefore, we can obtain a lower bound of the mutual information
\begin{equation}
\begin{aligned}
\label{eq:izy-lower}
I(\vec{Z},\vec{Y})
&\geq \sum_{\vec{y}, \vec{z}}q(\vec{y},\vec{z}) \log p(\vec{y}|\vec{z}) + H_q(\vec{Y})\\
&=\sum_{\vec{y}, \vec{z}, \vec{x}}q(\vec{x},\vec{y},\vec{z})\log p(\vec{y}|\vec{z})+ H_q(\vec{Y})\\ 
&\!\!\!=\!\!\!\sum_{\vec{y}, \vec{z}, \vec{x}}\!\!q(\vec{x}, \vec{y})q(\vec{z}|\vec{x}) \log p(\vec{y}|\vec{z})\!+\!H_q(\vec{Y}),
\end{aligned}
\end{equation}
where the last step uses $q(\vec{x},\vec{y},\vec{z}) = q(\vec{x})q(\vec{y}|\vec{x}) q(\vec{z}|\vec{x})$, which is a factorization based on the conditional dependency \footnote{$\vec{Y} \leftrightarrow \vec{X} \leftrightarrow \vec{Z}$: $\vec{Y}$ and $\vec{Z}$ are independent given $\vec{X}$.}. 

Since $q(\vec{X},\vec{Y},\vec{Z})$ is the approximation defined by ourselves, given a specific observation $(\vec{x}^{(i)},\vec{y}^{(i)})$, the empirical distribution $q(\vec{X}^{(i)},\vec{Y}^{(i)})$ is simply defined as a multiplication of two Delta functions
\begin{equation}
q(\vec{X}^{(i)}=\vec{x}^{(i)},\vec{Y}^{(i)}=\vec{y}^{(i)}) = \delta_{\vec{x}^{(i)}}(\vec{x})\cdot\delta_{\vec{y}^{(i)}}(\vec{y}).
\end{equation}
Then, \autoref{eq:izy-lower} with $\vec{X}^{(i)}$ and $\vec{Y}^{(i)}$ can be further simplified as
\begin{equation}
\begin{aligned}
\label{eq:vib-1}
I(\vec{Z};\vec{Y}^{(i)})
&\geq \sum_{\vec{z}}q(\vec{z}|\vec{x}^{(i)}) \log p(\vec{y}^{(i)}|\vec{z})\\
&=\expect{\log p(\vec{y}^{(i)}|\vec{z})}{q(\vec{z}|\vec{x}^{(i)})}
\end{aligned}
\end{equation}

\paragraph{The upper bound for $I(\vec{Z};\vec{X})$.}
\begin{eqnarray}
I(\vec{Z},\vec{X})
&=& \sum_{\vec{x}, \vec{z}} q(\vec{x},\vec{z})\log \frac{q(\vec{x},\vec{z})}{q(\vec{x}) q(\vec{z})}\nonumber\\
&=& \sum_{\vec{x}, \vec{z}} q(\vec{x},\vec{z})\log \frac{q(\vec{z}|\vec{x})}{q(\vec{z})}\nonumber\\
&=& \sum_{\vec{x}, \vec{z}} q(\vec{x},\vec{z})\log q(\vec{z}|\vec{x})\nonumber\\
& & - \sum_{\vec{x}, \vec{z}} q(\vec{x},\vec{z})\log q(\vec{z})
\end{eqnarray}
By replacing $q(\vec{z})$ with a prior distribution of $\vec{z}$, $p_0(\vec{z})$, we have
\begin{equation}
\sum_{\vec{x}, \vec{z}} q(\vec{x},\vec{z})\log q(\vec{z}) \geq \sum_{\vec{x}, \vec{z}} q(\vec{x},\vec{z})\log p_0(\vec{z}).
\end{equation}
Then we can obtain an upper bound of the mutual information
\begin{eqnarray}
\label{eq:vib-2}
I(\vec{Z},\vec{X})
&\leq& \sum_{\vec{x}, \vec{z}} q(\vec{x},\vec{z})\log q(\vec{z}|\vec{x})\nonumber\\
& & - \sum_{\vec{x}, \vec{z}}q(\vec{x},\vec{z})\log p_0(\vec{z})\nonumber\\
&=& \sum_{\vec{x}} q(\vec{x})\kldist{q(\vec{z}|\vec{x})}{p_0(\vec{z})}\nonumber\\
&=&  \expect{\kldist{q(\vec{z}|\vec{x})}{p_0(\vec{z})}}{q(\vec{x})}\nonumber\\
&=& \kldist{q(\vec{z}|\vec{x}^{(i)})}{p_0(\vec{z})}.
\end{eqnarray}

\section{Sampling with Gumbel-softmax Trick}
\label{sec:sampling}

Specifically, as $r_{x_t}$ has the probability $1-p_{x_t}$ and $p_{x_t}$ to take 0 or 1 respectively, we draw samples from a Gumbel(0, 1) distribution for each category $c\in\{0, 1\}$:
\begin{equation}
\label{eq:gumbel_softmax1}
s_{x_t}^{(c)}=-\log(-\log u),~u\sim \text{Uniform}(0, 1),
\end{equation}
and then apply a temperature-dependent softmax over the two categories with each dimension calculated as
\begin{equation}
\label{eq:gumbel_softmax2}
rs_{x_t}^{(c)}=\frac{\exp((w_{x_t}+s_{x_t}^{(c)})/\tau)}{\sum_{c}\exp((w_{x_t}+s_{x_t}^{(c)})/\tau)},
\end{equation}
where $w_{x_t}=\log(cp_{x_t}+(1-c)(1-p_{x_t}))$, and $\tau$ is a hyperparameter called the softmax temperature.

\section{Supplement of Experiment Setup}
\label{sec:sup_exp}

\paragraph{Data pre-processing.} We clean up the text by converting all characters to lowercase, removing extra whitespaces and special characters. We tokenize texts and remove low-frequency words to build vocab. We truncate or pad sentences to the same length for mini-batch during training. \autoref{tab:datasets_sup} shows pre-processing details on the datasets.

\begin{table}
	\centering
	\begin{tabular}{cccc}
		\toprule
		Datasets & $vocab$ & $threshold$ & $length$  \\
		\midrule
		IMDB & 29571 & 5 & 250  \\
		SST-1 & 17838 & 0 & 50  \\
		SST-2 & 16190 & 0 & 50 \\
		Yelp & 45674 & 10 & 150 \\
		AG News & 21838 & 5 & 50 \\
		TREC & 8026 & 0 & 15 \\
		Subj & 9965 & 1 & 25 \\
		\bottomrule
	\end{tabular}
	\caption{Pre-processing details on the datasets. $vocab$: vocab size; $threshold$: low-frequency threshold; $length$: mini-batch sentence length.}
	\label{tab:datasets_sup}
\end{table}

\paragraph{Model configurations.}
We implement the models in PyTorch 3.6. The number of parameters in the CNN, LSTM and BERT are 2652305, 2632405, 109486085 respectively. We tune hyperparameters manually for each model to achieve the best prediction accuracy. We experiment with different kernel numbers ($\{100, \cdots, 500\}$) for the CNN model, hidden states ($\{100, \cdots, 500\}$) for the LSTM model, and other hyperparameters, such as learning rate $lr \in \{1e-4, 1e-3,\cdots, 1\}$, clipping norm $clip \in \{1e-3, 1e-2, \cdots, 1, 5, 10\}$.

\paragraph{Implementation of L2X and IBA.}
\begin{itemize}
	\item The explanation framework of L2X \citep{chen2018learning} is a neural network which learns to generate importance scores $\vec{w}=[w_{1}, w_{2}, \cdots, w_{T}]$ for input features $\vec{x}=[\vec{x}_{1}, \vec{x}_{2}, \cdots, \vec{x}_{T}]$. The neural network is optimized by maximizing the mutual information between the selected important features and the model prediction, i.e. $I(\vec{x}_{S};y)$, where $\vec{x}_{S}$ contains a subset of features from $\vec{x}$. In our experiments, we adopt a single-layer feedforward neural network as the interpreter to generate importance scores for an input text, and multiply each word embedding with its importance score, $\vec{x}'=\vec{w}\odot \vec{x}$. The weighted word embedding matrix $\vec{x}'$ is sent to the rest of the model to produce an output $y'$. We optimize the interpreter network with the original model by minimizing the cross-entropy loss between the final output and the ground-truth label, $\mathcal{L}_{ce}(y_{t}; y')$.
	\item We adopt the Readout Bottleneck of IBA which utilizes a neural network to predict mask values $\vec{\lambda}=[\lambda_{1}, \lambda_{2}, \cdots, \lambda_{T}]$, where $\lambda_{t} \in [0, 1]$. The information of a feature $\vec{x}_{t}$ is restricted by adding noise, i.e. $\vec{z}_{t} = \lambda_{t}\vec{x}_{t}+(1-\lambda_{t})\vec{\epsilon}_{t}$, where $\vec{\epsilon}_{t}\sim\mathcal{N}(\mu_{\vec{x}_{t}}, \sigma_{\vec{x}_{t}}^{2})$. And $\vec{z}$ is learned by optimizing the objective function \autoref{eq:ob_func}. By assuming the variational approximation $q(\vec{z})$ as a Gaussian distribution, the mutual information can be calculated explicitly \citep{schulz2020restricting} . We still use a single-layer feedforward neural network as the Readout Bottleneck to generate continuous mask valuses $\vec{\lambda}$ and construct $\vec{z}$ for model to make predictions. The Readout Bottleneck is trained jointly with the original model by minimizing the sum of the cross-entropy loss $\mathcal{L}_{ce}(y_{t}; y)$ and an upper bound $\mathcal{L}_{I}=\expect{\kldist{p(\vec{z}|\vec{x})}{q(\vec{z})}}{\vec{x}}$ of the mutual information $I(\vec{Z};\vec{X})$. See \citet{schulz2020restricting} for the proof of the upper bound.
	
\end{itemize}

\section{Validation Performance and Average Runtime}
\label{sec:val_run}
The corresponding validation accuracy for each reported test accuracy is in \autoref{tab:val-acc}. The average runtime for each approach on each dataset is recorded in \autoref{tab:run-time}. All experiments were performed on a single NVidia GTX 1080 GPU.

\input{tables/tab-val-run}

\section{Results of Prediction Accuracy Reported in Previous Papers}
\label{sec:pre_results}
\autoref{tab:pre_acc} shows some results of prediction accuracy of base models reported in previous papers.

\input{tables/tab-pre-acc}

\section{Pearson Correlation between Word Frequency and \ourmethod Expectation Values}
\label{sec:p_corr}
\autoref{tab:p-corr} shows the Pearson correlation coefficients between word frequency and global word importance of \ourmethod-based models.

\input{tables/tab-p-corr}

%% file: tables/tab-val-run.tex
\begin{table*}[tbph]
	\centering
	\begin{tabular}{ccccccccc}
		\toprule
		Models & Methods & IMDB & SST-1 & SST-2 & Yelp & AG News & TREC & Subj \\
		\midrule
        \multirow{5}{*}{CNN} & CNN-base & 88.82  & 45.09  & 84.47 & 93.70  & 91.77 & 92.22 & 88.96 \\
		& CNN-$l2$ & 88.86 & 45.02  & 84.50  & 94.15  & 91.78 & 91.64 & 88.79 \\
		& CNN-L2X & 79.76 & 36.85  & 80.12 & 84.10  & 85.28 & 62.50 & 82.54 \\
		& CNN-IBA & 87.56 & 41.52  & 84.79 & 93.42  & 91.73 & 91.40 & 89.02 \\
		& CNN-\ourmethod & 88.98 & 44.78  & 85.56 & 94.34  & 91.78 & 92.60 & 91.60 \\
		\midrule
		\multirow{5}{*}{LSTM} & LSTM-base & 88.80 & 41.87  & 85.21 & 94.24  & 91.78 & 91.15 & 90.50 \\
		& LSTM-$l2$ & 88.82 & 41.90 & 85.22 & 94.21  & 91.80 & 91.17 & 90.46 \\
		& LSTM-L2X & 67.72 & 37.00  & 77.49 & 76.50  & 78.84 & 47.28 &82.07 \\
		& LSTM-IBA & 88.38 & 42.48  & 85.30 & 94.68  & 91.93 & 86.35 & 90.52 \\
		& LSTM-\ourmethod & 90.06 &  44.96 & 85.67 & 94.46  & 92.28 & 92.48 & 92.80 \\
		\midrule
		\multirow{5}{*}{BERT} & BERT-base & 85.16 & 50.77  & 92.66 & 96.78  & 93.88 & 95.13 & 95.30 \\
		& BERT-$l2$ & 85.05 & 49.32  & 92.65 & 96.80  & 93.69 & 95.36 & 94.87 \\
		& BERT-L2X & 65.34 &  38.17 & 72.68 &  83.00 & 80.67 & 91.32 & 84.59 \\
		& BERT-IBA & 85.60 & 51.02  & 91.74 & 96.50  & 93.24 & 95.28 & 96.10 \\
		& BERT-\ourmethod & 86.00 & 51.96  & 92.74 & 96.87  & 94.88 & 95.49 & 96.90 \\
		\bottomrule
	\end{tabular}
	\caption{Validation accuracy (\%) for each reported test accuracy.}
	\label{tab:val-acc}
\end{table*}

\begin{table*}[tbph]
	\centering
	\begin{tabular}{ccccccccc}
		\toprule
		Models & Methods & IMDB & SST-1 & SST-2 & Yelp & AG News & TREC & Subj \\
		\midrule
		\multirow{5}{*}{CNN} & CNN-base & 8.93  & 0.75  & 0.66 & 31.25  & 12.47 & 2.01 & 0.69 \\
		& CNN-$l2$ & 8.96 & 0.78  & 0.65 & 31.27  & 12.48 & 2.00 & 0.68 \\
		& CNN-L2X & 11.23 & 0.92  & 0.87 & 33.00  & 13.62 & 2.10 & 0.73 \\
		& CNN-IBA & 14.21 & 2.58  & 1.58 & 47.80  & 18.00 & 2.67 & 2.15 \\
		& CNN-\ourmethod & 12.29 & 1.17  & 1.21 & 38.59  & 16.16 & 2.58 & 1.12 \\
		\midrule
		\multirow{5}{*}{LSTM} & LSTM-base & 3.81 & 1.00  & 1.65 & 32.12  & 12.84 & 1.87 & 1.34 \\
		& LSTM-$l2$ & 3.83 & 1.01 & 1.67 & 32.13  & 12.84 & 1.89 & 1.34 \\
		& LSTM-L2X & 5.23 & 1.02  & 1.70 & 35.60  & 13.20 & 1.92 & 1.36 \\
		& LSTM-IBA & 8.39 & 1.20  & 1.85 & 38.10  & 15.54 & 2.01 & 1.90 \\
		& LSTM-\ourmethod & 5.07 & 1.07  & 1.73 & 36.70  & 13.39 & 1.96 & 1.47 \\
		\midrule
		\multirow{5}{*}{BERT} & BERT-base & 716.87 & 111.66  & 51.64 & 1846.12  & 1202.09 & 49.56 & 73.46 \\
		& BERT-$l2$ & 719.23 & 111.66  &  51.65 & 1846.12  & 1202.10 & 50.20 & 74.49 \\
		& BERT-L2X & 725.65 & 110.25  & 51.70 & 1845.30  & 1203.00 & 52.10 & 77.00 \\
		& BERT-IBA & 791.78 & 115.30  & 54.12 & 1885.02  & 1207.23 & 60.35 & 90.00 \\
		& BERT-\ourmethod & 774.35 & 110.79  & 53.64 & 1879.36  & 1205.35 & 57.48 & 85.85 \\
		\bottomrule
	\end{tabular}
	\caption{Average runtime (s/epoch) for each approach on each dataset.}
	\label{tab:run-time}
\end{table*}

%% file: tables/tab-pre-acc.tex
\begin{table*}
	\centering
	\begin{tabular}{ccccccccc}
		\toprule
		Models & Methods & IMDB & SST-1 & SST-2 & Yelp & AG News & TREC & Subj \\
		\midrule
        \multirow{2}{*}{CNN-base} & \citet{kim2014convolutional} & -  & 45.5  & 86.8  & -  & -  & 92.8 & 93.0 \\
        & \citet{wang2017combining} & - & - & - & - & 86.11 & 89.33 & - \\
		\midrule
		\multirow{4}{*}{LSTM-base} & \citet{du2019investigating} & - & 45.3  & 80.6  & -  & -  & 86.8 & 89.3 \\
		& \citet{zhou2015c} & - & 46.6 & 86.6 & - & - & - & - \\
		& \citet{zhang2015character} & - & - & - & 94.74 & 86.06 & - & - \\
		& \citet{liu2016recurrent} & 88.5 & 45.9 & 85.8 & - & - & - & - \\
		\midrule
		\multirow{2}{*}{BERT-base} & \citet{devlin2018bert} & - & -  & 93.5  & -  & - & - & - \\
		 & \citet{sun2019fine} & 94.6 & - & - & 97.72 & 94.75 & - & - \\
		\bottomrule
	\end{tabular}
	\caption{Results of prediction accuracy (\%) collected from previous papers.}
	\label{tab:pre_acc}
\end{table*}

%% file: tables/tab-p-corr.tex
\begin{table}
	\centering
	\begin{tabular}{llll}
		\toprule
		Datasets & CNN & LSTM & BERT \\
		\midrule
		IMDB & -0.045 & -0.018 & -0.016  \\
		SST-1 & -0.079 & -0.070 & 0.017  \\
		SST-2 & -0.067 & -0.068 & -0.023  \\
		Yelp & -0.014 & 0.004 & -0.038  \\
		AG News & -0.015 & -0.040 & -0.030  \\
		TREC & -0.018 & -0.026 & 0.028  \\
		Subj & -0.010 & -0.003  & 0.008 \\
		\bottomrule
	\end{tabular}
	\caption{Pearson correlation coefficients of \ourmethod-based models on the seven datasets.}
	\label{tab:p-corr}
\end{table}